\documentclass[a4paper,twoside]{article}

\usepackage{epsfig}
\usepackage{subcaption}
\usepackage{calc}
\usepackage{amssymb}
\usepackage{amstext}
\usepackage{amsmath}
\usepackage{amsthm}
\usepackage{multicol}
\usepackage{pslatex}
\usepackage{apalike}
\usepackage{hyperref} 
\usepackage{algorithmic}

\usepackage{algorithm2e}

\usepackage[bottom]{footmisc}
\usepackage{SCITEPRESS}    
\begin{document}

\title{Adaptive Structured Pruning of Convolutional Neural Networks \\for Time Series Classification}

\author{\authorname{Javidan Abdullayev\sup{1}\orcidAuthor{0009-0005-9142-8145}, Maxime Devanne\sup{1}\orcidAuthor{0000-0002-1458-3855}, Cyril	Meyer\sup{1}\orcidAuthor{0000-0001-7262-999X}, Ali Ismail-Fawaz\sup{1}\orcidAuthor{0000-0001-5385-3339}, Jonathan Weber\sup{1}\orcidAuthor{0000-0002-3694-4703}, Germain Forestier\sup{1, 2}\orcidAuthor{0000-0002-4960-7554}, }
\affiliation{\sup{1}IRIMAS, Université de Haute Alsace, Mulhouse, France}
\affiliation{\sup{2}DSAI, Monash University, Melbourne, Australia}
\email{\{javidan.abdullayev, maxime.devanne, cyril.meyer, ali-el-hadi.ismail-fawaz, jonathan.weber, germain.forestier \}@uha.fr}
}
\abstract{Deep learning models for Time Series Classification (TSC) have achieved strong predictive performance but their high computational and memory requirements often limit deployment on resource-constrained devices. 
While structured pruning can address these issues by removing redundant  filters, existing methods typically rely on manually tuned hyperparameters such as pruning ratios which limit scalability and generalization across datasets. 
In this work, we propose Dynamic Structured Pruning (DSP), a fully automatic, structured pruning framework for convolution-based TSC models. 
DSP introduces an instance-wise sparsity loss during training to induce channel-level sparsity, followed by a global activation analysis to identify and prune redundant filters without needing any predefined pruning ratio. 
This work tackles computational bottlenecks of deep TSC models for deployment on resource-constrained devices.
We validate DSP on 128 UCR datasets using two different deep state-of-the-art architectures: LITETime and  InceptionTime. 
Our approach achieves an average compression of 58\% for LITETime and 75\% for InceptionTime architectures while maintaining classification accuracy. 
Redundancy analyses confirm that DSP produces compact and informative representations, offering a practical path for scalable and efficient deep TSC deployment. }
\onecolumn \maketitle \normalsize \setcounter{footnote}{0} \vfill

\section{\uppercase{Introduction}}
\label{sec:introduction}
In recent years, Time Series Classification (TSC) has received significant attention from researchers due to its wide-ranging applications in domains such as human action recognition~\cite{nweke2018deep}, healthcare monitoring~\cite{rajkomar2018scalable}, fault detection~\cite{chadha2019time} and environmental event classification~\cite{mohaimenuzzaman2023environmental}.
Deep learning approaches, particularly convolutional architectures, have significantly advanced TSC task by achieving state-of-the-art (SOTA) performances across various benchmarks.
Despite these advances, many existing approaches overlook the relationship between task complexity and model complexity.
Consequently, large models are frequently applied uniformly across tasks of varying difficulty, resulting in unnecessary computational overhead and suboptimal deployment efficiency.

\begin{figure}[h]
    \centering
    \includegraphics[width=0.9\linewidth]{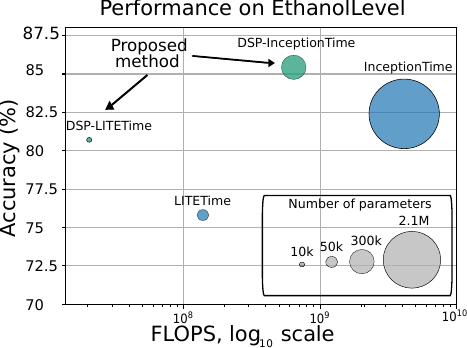}
    \caption{Accuracy versus FLOPS for baseline and pruned (DSP) models across LITETime and InceptionTime architectures on the EthanolLevel dataset. Marker size reflects the number of parameters (log-scaled). In this comparison, our DSP models achieve better test performance while significantly reducing computational cost and model size.}
    \label{fig:acc_vs_floops}
\end{figure}

To date, only a few works have explored model compression techniques such as knowledge distillation for TSC models~\cite{ay2022study,gong2022kdctime}.
Other compression methods including model pruning and quantization also offer a solution to reduce model size, computational cost and energy consumption without compromising performance. 
While unstructured pruning methods have been extensively studied, they often result in irregular sparsity patterns which require special hardware and software for efficient execution.
In contrast, structured pruning methods~\cite{luo2017thinet} remove entire filters, channels or blocks, resulting in compact and hardware-friendly models that are better suited for deployment on edge devices.
However structured pruning methods often rely on hyperparameters like manually set sparsity levels or pruning ratios which require careful tuning and may not generalize well across different datasets.
This hyperparameter dependency limits their practical applicability in diverse and large-scale TSC tasks where the optimal pruning ratio is unknown a priori.

In this work, we propose a method called Dynamic Structured Pruning (DSP), a fully automatic, structured pruning framework for deep learning-based TSC models.
Our DSP approach eliminates the need for manually defined pruning hyperparameters, offering a dynamic and adaptive solution to model compression.
To the best of our knowledge, this is the first work to introduce a fully end-to-end structured pruning framework that is data-agnostic and applicable to any task involving convolutional models.
Our approach first introduces an instance-wise sparsity loss during training to promote activation sparsity at the feature level.
Following training, DSP perform a global activation analysis across the dataset to identify and prune redundant  filters
without requiring any predefined pruning ratio.
We then retrain the pruned network to restore its performance.

We validate our DSP framework on 128 datasets from the UCR Archive~\cite{dau2019ucr} which spans a wide range of domains and sequence characteristics.
Our approach is evaluated on two different state-of-the-art deep learning architectures for TSC: LITETime~\cite{ismail2023lite} and InceptionTime~\cite{ismail2020inceptiontime}.
Experimental results demonstrate that DSP achieves significant compression, reducing model size by an average of 58\% for LITE and 75\% for Inception models while maintaining SOTA classification performance. 
As shown in Figure~\ref{fig:acc_vs_floops}, DSP models achieve higher accuracy compared to the original models with more than 85\% reduction in parameters on the Ethanol dataset.
The resulting models are compact, computationally efficient and well-suited for deployment on resource-constrained edge devices.

Our main contributions are as follows:
\begin{itemize}
    \item We introduce Dynamic Structured Pruning (DSP), a fully automatic, hyperparameter-free structured pruning framework for deep learning-based TSC models, eliminating the need for manually specified pruning ratios.
    \item We introduce an instance-wise sparsity training strategy combined with a global activation analysis to identify and remove redundant 
    filters without requiring any manual intervention.
    \item We conduct extensive experiments on 128 UCR datasets using two different state-of-the-art deep learning architectures, LITETime and InceptionTime, achieving average compression of 58\% and 75\% respectively while maintaining classification performance.
    Across 128 UCR datasets, our method compresses LITETime and InceptionTime by 58\% and 75\% on average while maintaining classiication accuracy, enabling efficient deployment on resource-constrained devices.
    
\end{itemize}

The rest of the paper is organized as follows: Section~\ref{sec:related_work} provides background information and reviews related works. 
Section~\ref{sec:method} presents our proposed pruning framework. 
Finally, Section~\ref{sec:conclusion} concludes the paper and outlines future research directions.

\begin{figure*}[h!]
    \centering
    \includegraphics[width=0.85\linewidth]{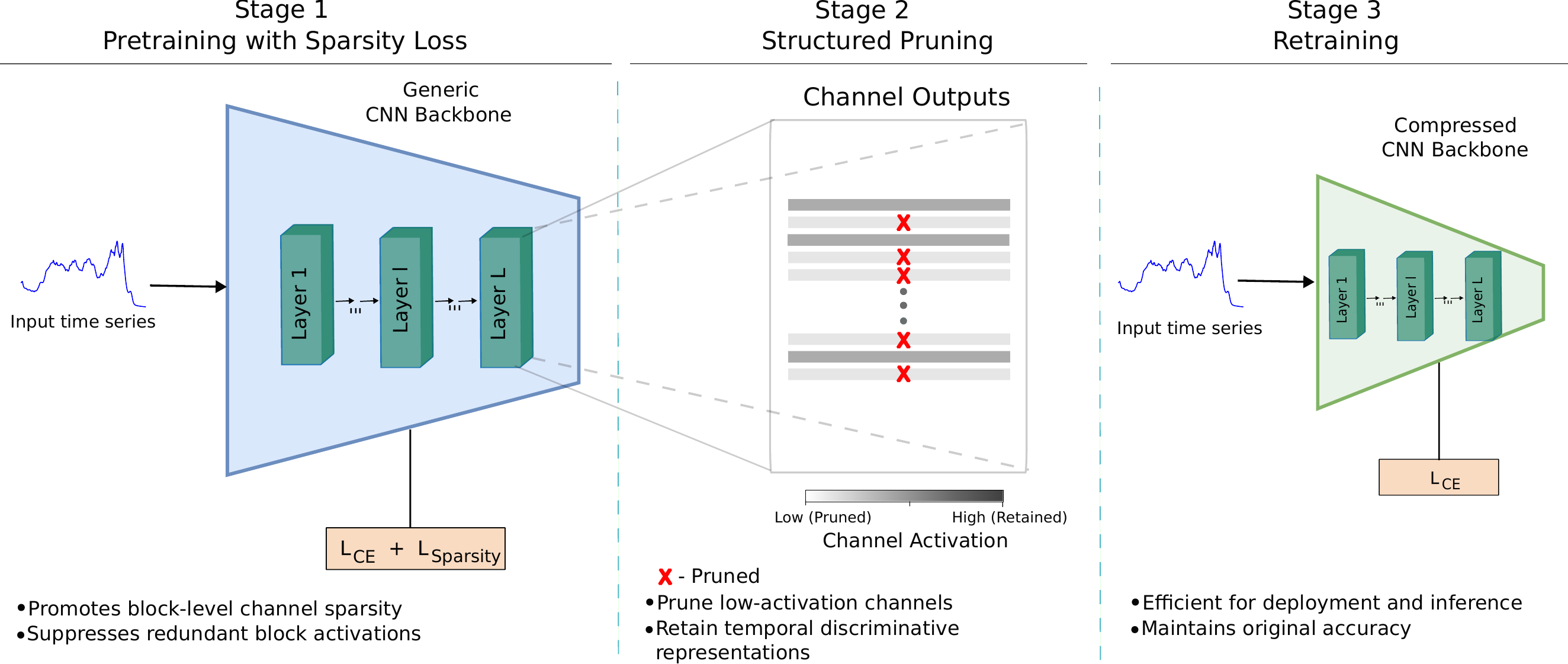}
    \caption{An overview of the proposed DSP framework.}
    \label{fig:overview}
\end{figure*}

\section{Background and Related Work}\label{sec:related_work}
Over the past decade, TSC field has transitioned from traditional feature-based techniques to deep learning models that automatically extract meaningful temporal patterns. 
Model compression techniques have become increasingly important in TSC due to the growing size and complexity of modern deep learning architectures. 
As these models become more computationally demanding, efficient deployment on resource-constrained devices relies on compression strategies that reduce redundancy without compromising performance.
This section introduces key definitions, reviews deep learning architectures for TSC and outlines major trends in model compression with a particular focus on pruning strategies.

\subsection{Definitions}
A univariate time series is defined as an ordered sequence of real-valued observations which represents evolution of a specific variable over time. 
Formally, a time series dataset is represented as $\mathcal{D} = {(\mathbf{X}_i, \mathbf{Y}i)}, i=1\dots N$, where $N$ is the number of samples, $\mathbf{X}_i \in \mathbb{R}^{T}$ denotes a univariate time series of length $T$ and $\mathbf{Y}_i$ is its corresponding class label.
The goal of the TSC task is to learn a function $f: \mathbf{X} \rightarrow \mathbf{Y}$ that maps each input sequence to its correct class label by capturing discriminative temporal patterns.

\subsection{Deep Learning for Time Series Classification}
Some traditional TSC approaches, particularly those based on classifiers such as Random Forests~\cite{lucas2019proximity} or SVMs~\cite{bagheri2016support}, rely heavily on a feature engineering step.
Because feature extraction and classification are handled separately, these pipelines often introduced information loss and added complexity.



Deep learning models were adopted to address these issues by learning representations and classifiers jointly from raw time series.
Early models like MLPs~\cite{ismail2019deep} were limited in capturing temporal dependencies due to their fully connected structure. 
Introduction of 1D Convolutional Neural Networks (CNNs), particularly Fully Convolutional Network (FCN)~\cite{wang2017time}, marked a breakthrough by learning discriminative features directly from raw inputs. 
ResNet~\cite{wang2017time} extended this with residual connections for improved training stability.

InceptionTime~\cite{ismail2020inceptiontime}, inspired by Google's Inception architecture~\cite{szegedy2017inception}, introduced modules with multiple kernel sizes to capture features at different temporal scales. 
Later, H-Inception~\cite{ismail2022deep} enhanced this design with custom filters to improve performance.
LITE~\cite{ismail2023lite}, proposed as a lightweight alternative to InceptionTime, achieves similar accuracy with reduced complexity. 
LITE uses custom, dilated, multiplexed and depthwise separable convolutions to balance performance and efficiency.
Recently, CoCaLite~\cite{badi2024cocalite} extended this line of work by incorporating ensemble strategies to further improve efficiency.

Alongside these deep learning approaches, several classical machine learning methods continue to remain competitive.
ROCKET (Random Convolutional Kernels)~\cite{dempster2020rocket} is an efficient method that uses a large set of random convolutional kernels to extract features from time series, followed by a simple linear classifier. MultiROCKET~\cite{tan2022multirocket} extends this by incorporating multiple kernel sizes and it has demonstrated strong performance across a range of TSC tasks. 
Hive-COTE~\cite{middlehurst2021hive} is an ensemble method that combines many classical and deep learning techniques to create a robust model for TSC.

\subsection{Model Compression Techniques}
Model compression techniques such as pruning, quantization and knowledge distillation have been extensively studied to reduce the computational cost and memory footprint of CNNs. 
Early works demonstrated that network pruning could effectively reduce model complexity and mitigate overfitting~\cite{lecun1989optimal,hanson1988comparing,hassibi1992second}.
A data-free pruning approach for dense layers was introduced by~\cite{srinivas2015data} where neurons with similar vectors are merged to reduce redundancy.
Fine-grained pruning reduces CNN size by removing individual weights while preserving accuracy. 
Sensitivity-based methods~\cite{lecun1989optimal,hassibi1992second} use second-order derivatives to prune low-impact weights. 
Magnitude-based pruning~\cite{han2015learning} removes weights with the smallest values, assuming they matter least.
However, these sparsity-based methods produce irregular weight patterns that require specialized hardware systems~\cite{han2016eie,chen2018escoin} to achieve speedup.

This motivates the exploration of structured pruning approaches that can offer more predictable and hardware-friendly sparsity patterns.
Heuristic metrics have been proposed for this purpose, such as using the $\ell_1$ norm of convolutional kernels as an indicator of filter importance~\cite{filters2016pruning} where filters with smaller norms are pruned.
In~\cite{hu2016network}, authors proposed Network Trimming, a data-driven pruning method that removes filters with high zero activation rates based on validation data.
In addition, some works frame  pruning as a feature selection problem using statistical tools~\cite{he2017channel,luo2017thinet,molchanov2016pruning}. 
The works~\cite{luo2017thinet,he2017channel} apply Lasso regression to identify filters that minimize the reconstruction error of the feature maps of the next layer.
A more principled approach employs the Taylor expansion to estimate the contribution of each filter to the final loss, allowing the removal of filters with minimal impact~\cite{molchanov2016pruning}. 
Another research direction explored instance-wise sparsity to speed up  deep learning model inference~\cite{mohaimenuzzaman2023environmental}.
In~\cite{liu2019learning}, authors proposed a method that learns a dynamic sparsity mask for each input instance by optimizing a sampling-based objective.

However, most of the methods mentioned above have primarily been applied to computer vision tasks and have not been adapted or specifically evaluated for TSC tasks.
To date, only a few works have explored model compression techniques such as knowledge distillation for TSC models~\cite{ay2022study,gong2022kdctime}.
This highlights the gap in applying and evaluating these techniques in TSC problems.

While most of the above methods rely on fixed sparsity levels and make pruning decisions based on per-sample activations, they often overlook that a filter inactive for a certain input may still be highly informative for others. 
This can lead to the premature removal of useful filters. 
DSP addresses this limitation by leveraging instance-wise sparsity loss during training, followed by a global activation-based pruning strategy that evaluates filter importance across the entire dataset. 
This enables dynamic identification of redundant filters without requiring prior knowledge of pruning ratios. 
As a result, DSP supports scalable and adaptive compression tailored to dataset complexity and provides a practical solution for deploying efficient deep models in TSC.

\section{Proposed method}\label{sec:method}

\subsection{Overall approach}
\label{subsec:approach}
We propose Dynamic Structured Pruning (DSP), a fully end-to-end structured pruning framework designed for deep learning-based TSC models.
Unlike traditional methods that require manual tuning of hyperparameters such as pruning ratios, DSP eliminates the need for predefined pruning configurations and automatically identifies redundant
filters
based on dataset complexity.
Our framework, specially designed for convolutional TSC models, combines three key components: \textit{instance-wise sparsity training},  \textit{global feature pruning} and \textit{adaptive retraining} as shown in Figure~\ref{fig:overview}. 
The figure provides an overview of the proposed method, from pretraining with sparsity loss to pruning and final retraining.
A detailed step-by-step description of the framework is summarized in Algorithm~\ref{alg:pruning_framework}. 
Further explanations of each component are provided in the following subsections.

\subsection{Pretraining with Sparsity Loss}
\label{subsec:sparsity_loss}
\begin{figure}[h]
    \centering
    \includegraphics[width=0.75\linewidth]{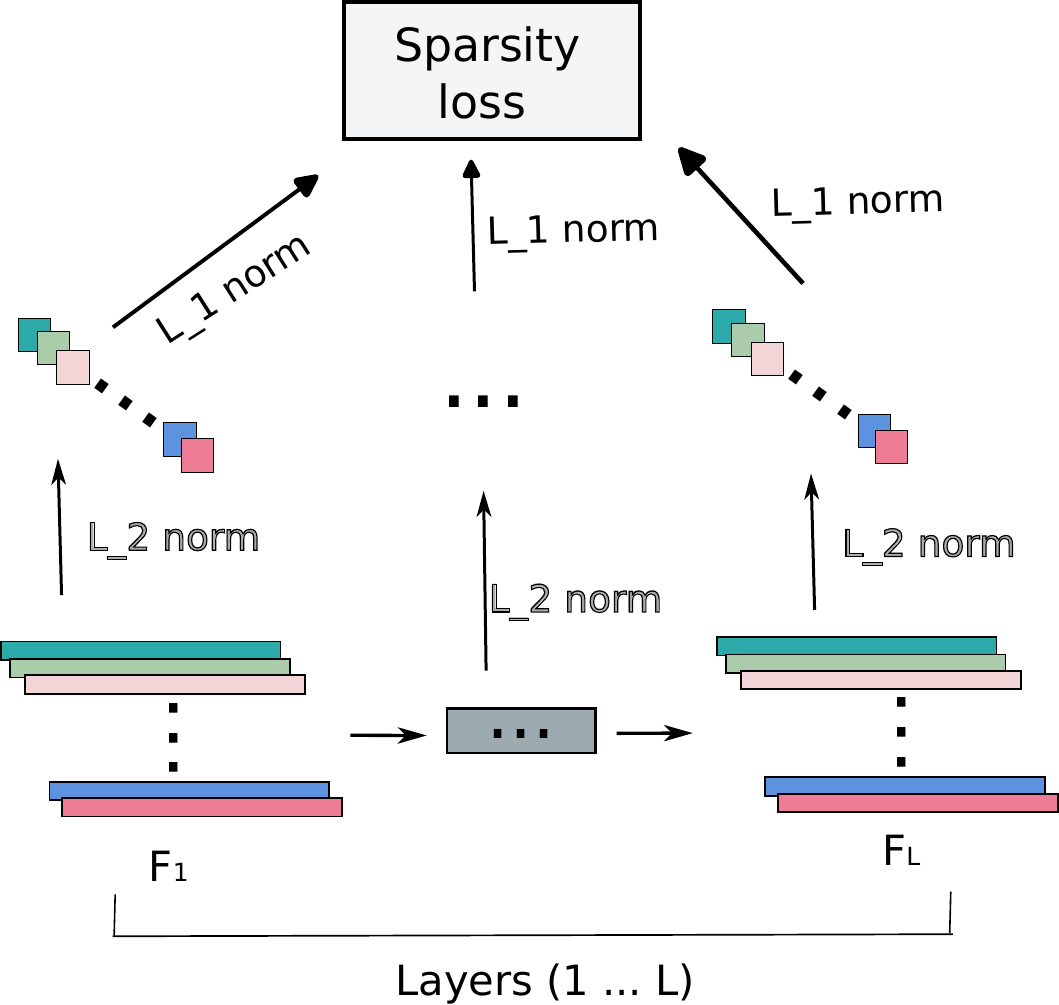}
    \caption{Illustration of sparsity-based training using instance-wise sparsity loss to induce channel sparsity 
    .
    }
    \label{fig:sparsity-based-training}
\end{figure}

First stage of DSP trains the model with an auxiliary sparsity constraint to induce structured sparsity in the learned representations.
Our instance-wise sparsity loss operates directly on the activation space unlike conventional weight regularizers that penalize individual parameters.
This design allows the model to suppress uninformative channel responses while preserving task-relevant dynamics.
The detailed illustration of the sparsity loss is shown in Figure~\ref{fig:sparsity-based-training}.

The sparsity loss in DSP operates hierarchically on the 3D feature tensor $F \in \mathbb{R}^{B \times C \times T}$ (batch $\times$ channels $\times$ time steps) as illustrated in Equation~\ref{eq:sparsity_loss}.

\begin{equation}
L_{\text{sparsity}} = \underbrace{\sum_{i=1}^{B} \sum_{j=1}^{C} \sqrt{\sum_{k=1}^{T} F[i,j,k]^2}}_{\text{Channel-wise $L_{2,1}$ penalty}}
\label{eq:sparsity_loss}
\end{equation}

The sparsity loss adopts a hierarchical $L_{2} + L_{1}$ formulation. 
For each input instance, an $L_{2}$-norm is first computed across the temporal activations within each channel which provides an energy measure that reflects the overall strength of the given channel response.
Then, an $L_{1}$-norm is applied across these channel energies to enforce global sparsity by encouraging entire channels with consistently low activation magnitudes to become inactive.

This hierarchical regularization enables DSP to perform superior feature selection compared to standard weight decay, as it specifically targets channel redundancy while maintaining the temporal integrity of discriminative patterns. 
The resulting sparse features provide an optimized starting point for the pruning phase of DSP.
The training loss in DSP is shown in the Equation~\ref{eq:train_loss}.
\begin{equation}
    \mathcal{L}_{\text{train}} = \mathcal{L}_{\text{CE}} + \lambda \cdot \mathcal{L}_{\text{sparsity}}
    \label{eq:train_loss}
\end{equation}

$\mathcal{L}_{\text{CE}}$ is the cross-entropy loss which ensures accurate classification and $\lambda$ is a weight parameter that balances the impact of the sparsity loss.

\subsection{Pruning Strategy}
\label{subsec:pruning}
After pretraining, DSP applies a data-driven pruning strategy that removes globally redundant
filters
while preserving discriminative temporal patterns.
Given the feature map $\mathbf{F} \in \mathbb{R}^{N \times C \times T}$ where $N$, $C$ and $T$ represent the total number of training samples, channels and time steps respectively.
DSP computes the activation strength of each channel using temporal $L_2$-normalizations
as shown in Equation~\ref{eq:activation_strength}.
\begin{equation}
s_{ij} = |\mathbf{F}[i,j,:]|_{2} = \sqrt{\sum_{k=1}^{T} |F[i,j,k]|^2}
\label{eq:activation_strength}
\end{equation}
where $i$, $j$ and $k$ represent the sample index, channel index and time step index respectively.

For each sample $i$, DSP determines an adaptive activation threshold $\tau_i$ as the mean activation strength across all channels:

\begin{equation}
\tau_i = \frac{1}{C}\sum_{j=1}^C s_{ij}.
\end{equation}

A binary pruning mask $\mathbf{M} \in {0,1}^{N \times C}$ is then constructed:

\begin{equation}
    M[i,j] = \begin{cases} 
        1 & \text{if } s_{ij} \geq \tau_i \quad \text{(active filter)} \\
        0 & \text{otherwise} \quad \text{(inactive filter)}
    \end{cases}
    \label{eq:pruning_mask}
\end{equation}

\begin{figure}[h]
    \centering
    \includegraphics[width=0.6\linewidth]{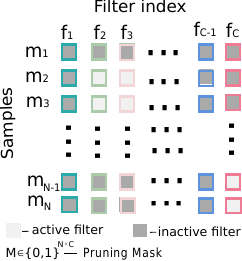}
    \caption{Pruning strategy that leverages feature activations across the dataset samples. Here, 
$f_{c}$ refers to the output activation of the $c^{th}$ channel for a given instance and $m_{n}$ represents one of the N input instances in the dataset.}
    \label{fig:pruning_strategy}
\end{figure} 

Basically, in DSP, the pruning decision is made by comparing activation of each channel to the mean activation value within the same instance.
Specifically, for a given input sample, we compute the average activation across all channels in a block and treat it as a reference level of information contribution.
Channels whose activation energy falls below this instance-wise mean are considered weakly responsive and are selected for pruning.
The underlying assumption is that, for a given input, discriminative channels respond more strongly than redundant ones which remain close to the mean background activity.
Using the instance-wise mean provides an adaptive, data-dependent threshold that reflects the internal activation dynamics of the model without relying on manually chosen hyper-parameters or global sparsity ratios.

Figure~\ref{fig:pruning_strategy} visually summarizes this pruning strategy.
The key innovation in DSP lies in the global consensus pruning criterion: a filter $j$ is pruned only if it is inactive for all samples $\sum_{i=1}^N M[i,j] = 0$.

DSP offers three distinct advantages over existing methods. 
It is immune to class-specific bias through complete activation-history analysis and preserves temporally sparse yet discriminative features via $L1$ normalization.
In addition, it employs a single-step pruning process that avoids the optimization instability of iterative pruning–retraining cycles observed in prior works~\cite{liu2019learning,mohaimenuzzaman2023environmental}

\subsection{Adaptive Retraining}
\label{subsec:retraining}
Following pruning, DSP retrains the model to adapt to the reduced architecture and maintain efficiency.
We explore two strategies, fine-tuning and scratch training. 
Fine-tuning updates the remaining parameters based on the original objective.
In contrast, scratch training reinitialize the pruned model weights and trains it from scratch, allowing the model to learn from the simplified architecture.

\begin{algorithm}[htbp]
    \footnotesize 
    \caption{Dynamic Structured Pruning (DSP) Framework}
    \label{alg:pruning_framework}
    \begin{algorithmic}[1]
        \STATE \textbf{Input:} Dataset $\mathcal{D}$, Base model $f_{\text{base}}$, Losses $\mathcal{L}_{\text{CE}}$, $\mathcal{L}_{\text{sparsity}}$
        \STATE \textbf{Output:} Pruned model $f_{\text{pruned}}$
        \STATE \textbf{Step 1: Pretraining with Sparsity Loss}
        \STATE Initialize $f_{\text{base}}$
        \STATE Train the model with $\mathcal{L} = \mathcal{L}_{\text{CE}} + \mathcal{L}_{\text{sparsity}}$
        \STATE Sparsity loss: $L_{\text{sparsity}} = \sum_{i=1}^{B} \sum_{j=1}^{C} \sqrt{\sum_{k=1}^{T} F[i,j,k]^2}$
        \STATE Obtain sparsity-optimized feature maps
        \STATE \textbf{Step 2: Global Activation Analysis and Pruning}
        \STATE Calculate $s_{ij}$ using $L_2$-norm: $s_{ij} = \sqrt{\sum_{k=1}^{T} |F[i,j,k]|^2}$
        \STATE Compute threshold: $\tau_i = \frac{1}{C}\sum_{j=1}^C s_{ij}$
        \STATE Create pruning mask $M[i,j]$:
        \[
        M[i,j] = \begin{cases} 
        1 & \text{if } s_{ij} \geq \tau_i \\
        0 & \text{otherwise}
        \end{cases}
        \]
        \STATE Prune redundant channels: $\sum_{i=1}^N M[i,j] = 0$
        \STATE \textbf{Step 3: Retraining the Pruned Model}
        \STATE Retrain pruned model with cross-entropy loss
    \end{algorithmic}
\end{algorithm}
\section{Experimental Evaluation}\label{sec:experimental_evaluation}
\subsection{Experimental Setup}
\subsubsection{Data}
To evaluate the performance of our proposed approach, we utilized 128 datasets from the UCR Time Series Classification Archive~\cite{dau2019ucr}
which spans diverse domains. 
These datasets exhibit significant variations in key characteristics, including time series length, number of classes (ranging from 2 to 60) and training set sizes. 
For a fair and consistent comparison with SOTA methods, we followed the standard train-test splits provided by the archive. 
Additionally, all time series were z-normalized to ensure consistent scaling across datasets.

\subsubsection{Compared Model Variants}
To evaluate the impact of pruning and retraining, we begin by comparing five model variants: the original uncompressed model (\textit{Base}), the sparsity-pretrained model before pruning (\textit{Pretrained}), the sparsity-pretrained model after pruning (\textit{Pruned}), the pruned model with fine-tuning (\textit{Finetuned}) and the pruned model retrained from scratch initialization (\textit{Scratch-trained}). 
These variants allow us to assess the effect of each stage of the pruning and retraining pipeline on model performance.
The specific naming conventions for these models are summarized in Table~\ref{tab:naming_convensions}. 

We perform five independent training runs for baseline and each configuration of DSP models.
The final performance metrics are derived from the ensemble accuracy across the five runs.
All comparisons in this work are based on these ensemble results as ensembles of five LITE (LITETime) and Inception (InceptionTime) models achieve SOTA performance on UCR datasets.

\begin{table}[h!]
    \setlength{\tabcolsep}{9pt}
    \centering
    \scriptsize
    \caption{Naming Conventions for Different Models}
    \begin{tabular}{ll}
        \hline
        \textbf{Model Type} & \textbf{Name} \\
        \hline
        Original Model & Base\\
        Pretrained Model & Pretrained\\
        Pretrained + Pruned Model & Pruned \\
        Pretrained + Pruned + Fine-Tuned Model & Finetuned \\
        Pretrained + Pruned + Scratch-Trained Model & Scratch trained \\
        \hline
    \end{tabular}
    \label{tab:naming_convensions}
\end{table}

\begin{figure*}[t]
    \centering
    \hspace{0.8cm}
    \begin{subfigure}[b]{0.7\linewidth}
        \includegraphics[width=\linewidth]{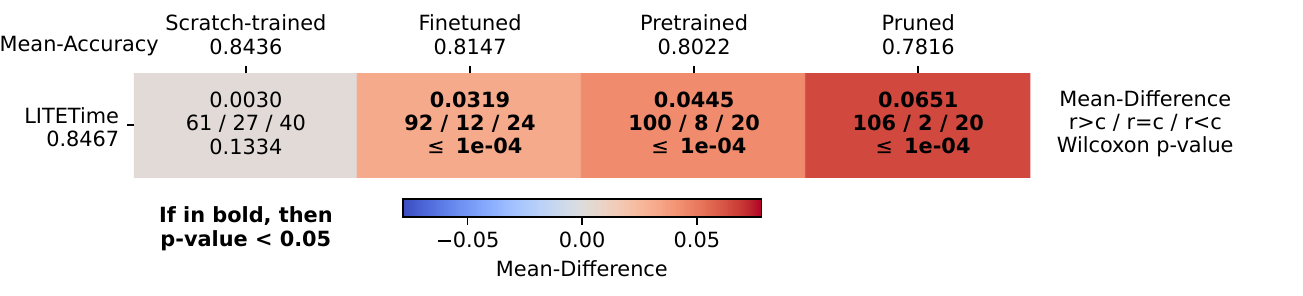}
        \caption{LITETime variants.}
        \label{fig:mcm_baselines_lite}
    \end{subfigure}
    
    \vspace{\floatsep} 
    
    \begin{subfigure}[b]{0.7\linewidth}
        \includegraphics[width=\linewidth]{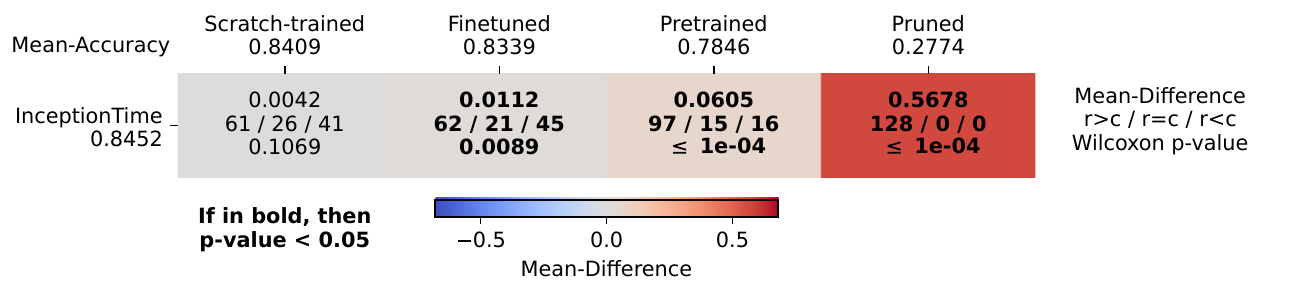}
        \caption{InceptionTime variants.}
        \label{fig:mcm_baselines_inception}
    \end{subfigure}
    
    \caption{Comparison of \textit{Base}, \textit{Pretrained}, \textit{Pruned}, \textit{Finetuned}, and \textit{Scratch-trained} models for LITE and Inception architectures.}
    \label{fig:mcm_baselines}
\end{figure*}

\subsubsection{Comparison Protocol}
We evaluate classification performance using accuracy as the primary metric. 
Our analysis employs the Multi-Comparison Matrix (MCM)~\cite{ismail2023approach} which provides three key advantages over traditional ranking methods: (1) Mean-Accuracy for overall performance assessment, (2) Mean-Difference for pairwise classifier comparisons (positive values indicate row outperforms column) and (3) Win/Tie/Loss counts across datasets.
Statistical significance of the comparisons is assessed using the Wilcoxon signed-rank test~\cite{wilcoxon1992individual} with a significance level of $p < 0.05$. 
Results that meet this criterion are highlighted in bold.
This approach ensures robust evaluation of our proposed approach against baseline methods.

\subsubsection{Implementation details}
We used both LITETime~\cite{ismail2023lite} and InceptionTime~\cite{ismail2020inceptiontime} models as the base architectures, following the original configurations. 
Training was performed using the Adam optimizer with an initial learning rate of 0.001 along with a learning rate scheduler (reduction factor = 0.5, patience = 50). 
Each model was trained for 1500 epochs using a batch size of 64.
All experiments were run on a system with an NVIDIA RTX 4090 GPU (24GB memory) 
running Ubuntu 22.04, implemented in PyTorch 2.5.1 and Python 3.12. 
The source code is publicly available \url{https://github.com/MSD-IRIMAS/Pruning4TSC}.

\subsection{Overall performance on UCR Archive}
\subsubsection{Impact of Pruning and Retraining}
We present comparative results across the five model configurations listed in Table~\ref{tab:naming_convensions} using an MCM as shown in Figure~\ref{fig:mcm_baselines}.
Across both architectures, \textit{Pruned} model shows a clear drop in performance.
This drop can be attributed to the high average pruning ratios, 58.2\% for LITETime and 74.6\% for InceptionTime models.
The \textit{Pretrained} models perform better than the \textit{Pruned} versions which is intuitive since they retain the full representational capacity of the original network.
\textit{Finetuned} models recover some of the lost performance but in both architectures they still remain statistically inferior to the baseline models.
In contrast, \textit{Scratch-trained} models match baseline performance and are statistically non-different in both MCM matrices.
These results underscore three key findings: (1) scratch training is essential for fully recovering after pruning, (2) our pruning strategy enables significant compression without requiring hyperparameter tuning and (3) the proposed method generalizes well across different architectures.
These results on scratch-training align with previous findings in time series model compression~\cite{mohaimenuzzaman2023environmental}.
The scratch-trained pruned models are referred to as DSP-LITETime and DSP-InceptionTime for LITETime and InceptionTime architectures respectively.

\begin{figure}
    \centering
    \begin{subfigure}[b]{0.83\linewidth}
        \includegraphics[width=\linewidth]{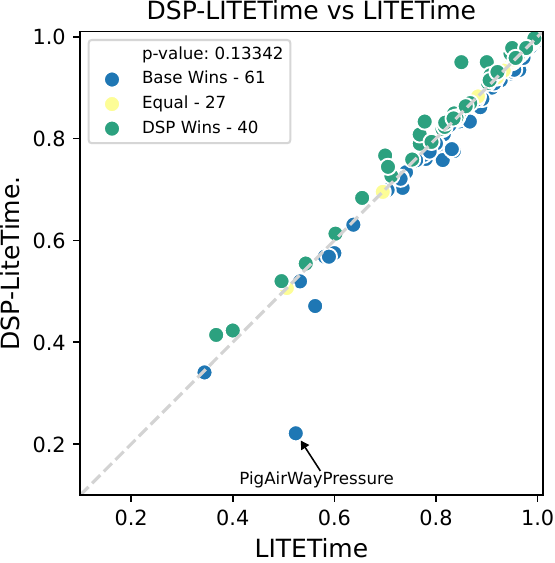} 
            \caption{\centering{Performance comparison for LITETime.}}
        \label{fig:scratch-trained-lite}
        \vspace{1em}    
        
    \end{subfigure}

    \begin{subfigure}[b]{0.83\linewidth}
        \includegraphics[width=\linewidth]{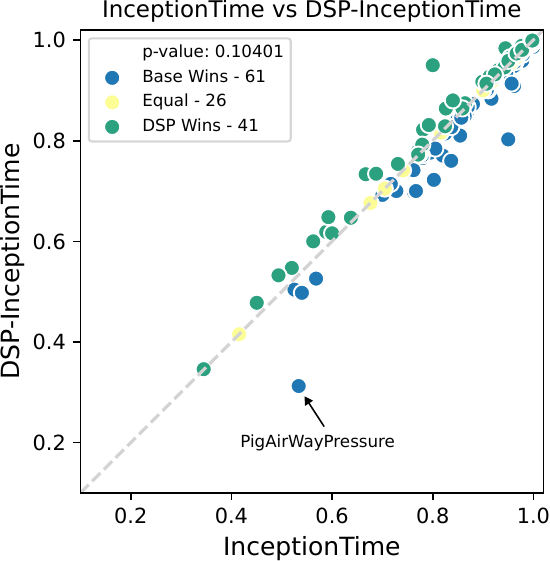}
        \caption{\centering{Performance comparison for InceptionTime}}
        \label{fig:scratch-trained-inception}
    \end{subfigure}

    \caption{\centering{Comparison of DSP models with baselines.}}
    \label{fig:combined-scratch-trained}
\end{figure}

\begin{figure*}
    \centering
    \includegraphics[width=0.9\linewidth]{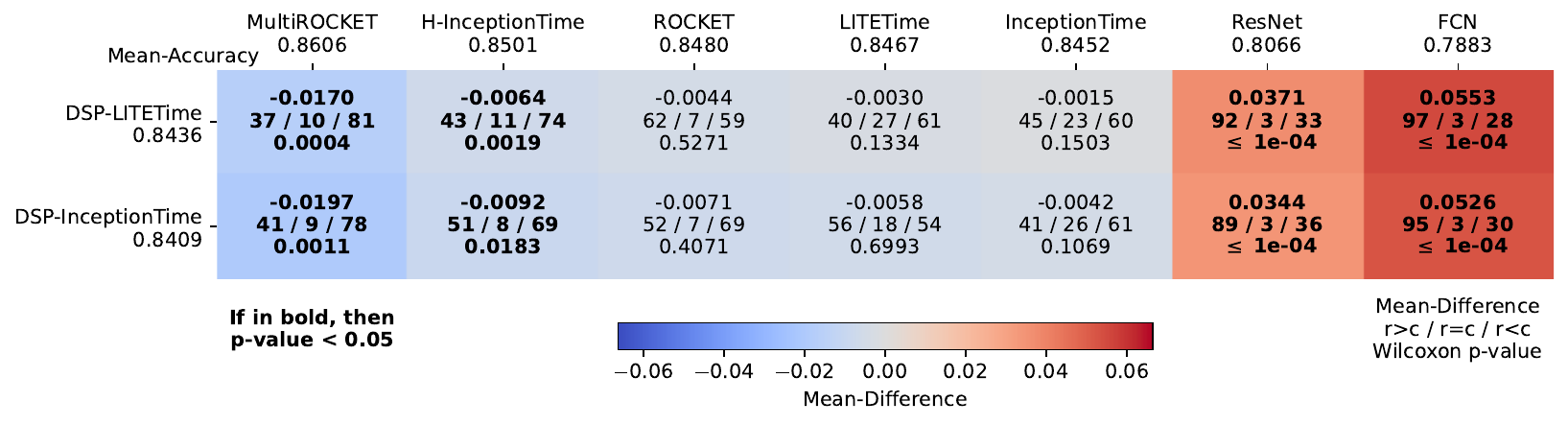}
    \caption{\centering{Comparisons of our DSP method with state-of-the-art approaches.
    }}
    \label{fig:mcm_sota}
\end{figure*}   

\subsubsection{Comparison with Baselines}
To further investigate the results we analyze a one-versus-one performance comparison between DSP models and their corresponding baselines, as illustrated in Figure~\ref{fig:scratch-trained-lite} and~\ref{fig:scratch-trained-inception}. 
The pairwise comparison reveals interesting insights regarding our pruning approach. 
DSP models achieve performance comparable to their counterparts on both architectures. 
The distribution of points shows strong clustering along the diagonal. 
Additionally, from the figure we can observe that DSP models tend to improve more on datasets where the baseline already achieves higher accuracy. 
This can be explained by the fact that when the task is relatively simple for the base model, DSP can effectively act as a regularizer to enhance overall performance. 
However, these figures also show that on the PigAirwayPressure dataset, DSP models perform worse than the baselines with a large margin.
The dataset contains 102 training samples only with 52 classes which means limited data per class and makes it hard for pruning to preserve all important features.
Both architectures exhibit approximately 20\% of datasets with tied performance which reflects the effectiveness of the method in preserving accuracy.

Additionally, Table~\ref{tab:model_comparison} shows complexity comparison between the proposed DSP models and corresponding baselines over the 128 UCR datasets.
From the results we can observe for both architectures DSP models significantly reduces number of paramaters, FLoating-point Operations Per Second (FLOPS) and memory size which enhances their efficiency for deployment on edge devices, robots or microcontrollers.
From these results we can conclude that DSP models maintain performance comparable to their baseline counterparts.
They offer substantial reductions in computational cost and memory usage which makes them suitable for resource-constrained environments.

\begin{table}[h]
    \setlength{\tabcolsep}{10pt}
    \centering
    \scriptsize
    \caption{Complexity comparison between baseline and DSP models.}
    \begin{tabular}{llll}
        \hline
        Model & \# Parameters & FLOPS & Memeory \\
        \hline
            LITETime        & 50,511      & 57M    & 0.2MB    \\
            DSP-LITETime    & 21,298      & 22M    & 0.09MB   \\
            InceptionTime   & 2,106,593   & 2265M  & 8.04MB   \\
            DSP-InceptionTime & 536,743    & 443M   & 2.05MB   \\
        \hline
    \end{tabular}
    \label{tab:model_comparison}
\end{table}

\subsubsection{Comparison with state-of-the-art}

We evaluate our proposed DSP models
against several state-of-the-art methods using MCM presented in Figure~\ref{fig:mcm_sota}.
From these results we can notice that MultiROCKET and H-InceptionTime outperform DSP models with statitically significant $p$-value but it is important to note that these methods also surpass their respective baselines (LITETime, InceptionTime)~\cite{ismail2022deep,ismail2020inceptiontime}.
It indicates that the performance gap is due to the architectural differences rather than our pruning approach.
Another state-of-the-art approach, ROCKET shows no statistically significant difference compared to DSP models.
As we observed from the Figure~\ref{fig:combined-scratch-trained}, pruning with DSP does not \textit{downgrade} performance compared to the uncompressed LITETime and InceptionTime. 
Notably, DSP models statistically outperform traditional architectures like ResNet and FCN.
These results show that our approach maintains the competitive position of its base architectures while delivering significant reductions in parameters, FLOPS and memory usage.

\subsection{Analysis of ratio reduction}
\begin{figure}
    \centering
    \begin{subfigure}[b]{0.8\linewidth}
        \includegraphics[width=\linewidth]{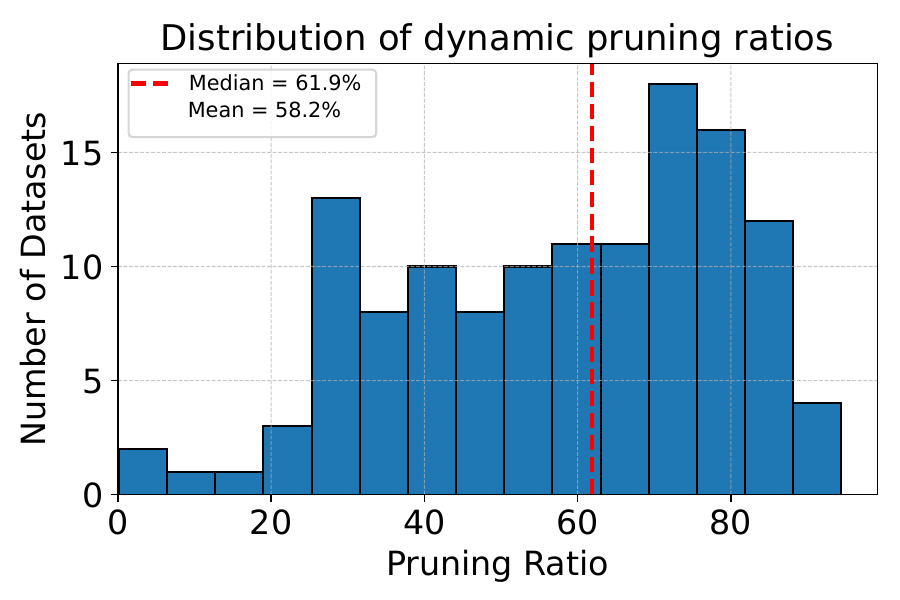}
        \caption{\centering{LITETime model.}}
        \label{fig:dynamic_pruning_hist}
    \end{subfigure}
    
    
    \begin{subfigure}[b]{0.8\linewidth}
        \includegraphics[width=\linewidth]{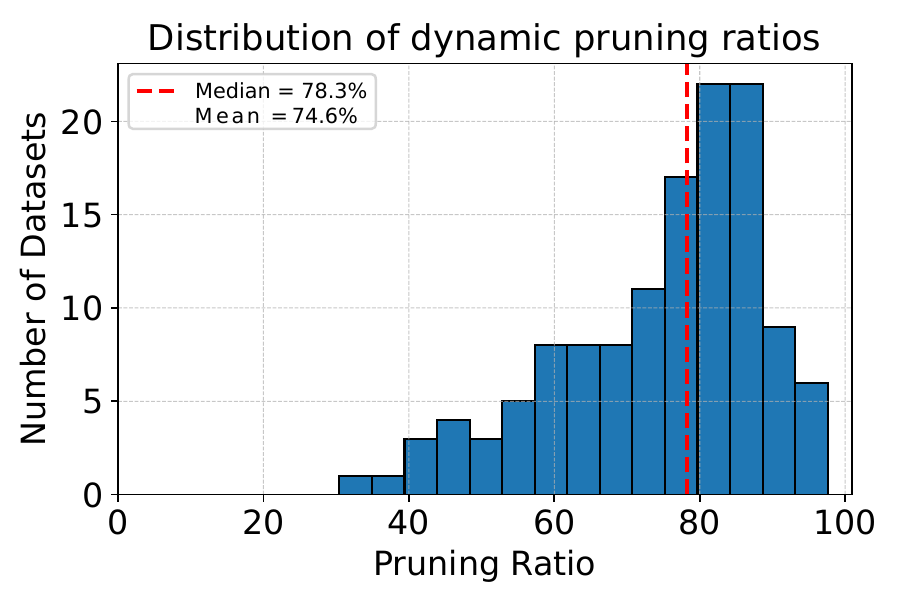}
        \caption{\centering{InceptionTime model.}}
        \label{fig:dynamic_pruning_hist_inc}
    \end{subfigure}
    
    \caption{\centering{Distribution of pruning ratios achieved by the DSP method across 128 UCR datasets.}}
    \label{fig:pruning ratios}
\end{figure}

In this part, we analyze the distribution of pruning ratios achieved by our DSP approach on both LITETime and InceptionTime 
architectures across 128 UCR datasets as shown in Figure~\ref{fig:pruning ratios}. 
As illustrated in Figure~\ref{fig:dynamic_pruning_hist}, pruning ratios for the LITETime model exhibit significant variation, ranging from as low as 0.01\% to more than 90\% with an average ratio of 58.2\% and a median of 61.9\%.
This distribution highlights the adaptive nature of our method, selectively applying pruning based on dataset complexity.
In the cases where the pruning ratio is close to zero can be explained by the fact that LITE is already a relatively lightweight architecture which means that our DSP approach retains most of the model for difficult datasets.
For the InceptionTime model (Figure~\ref{fig:dynamic_pruning_hist_inc}), pruning ratios are even more aggressive, ranging from 30.4\% up to a substantial 97.6\%. 
The average pruning ratio is significantly higher at 74.6\% with a median of 78.3\%. 
This is logical considering that the Inception architecture is much complex compared to LITE which means providing greater potential for pruning.
These results confirm that our DSP method dynamically adapts the pruning ratio based on dataset complexity and the underlying model architecture.
As a result, it achieves an effective trade-off between reducing model size and maintaining classification performance for different architectures.

\subsection{Dynamic vs Static Pruning}
\begin{figure*}
    \centering
     \hspace{3cm} 
    \includegraphics[width=0.63\linewidth]{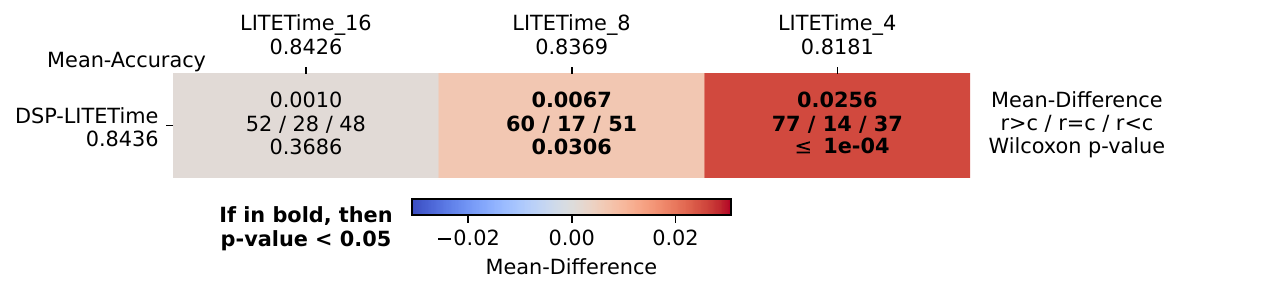}
    \caption{\centering{Comparison of our DSP approach with statically pruned models.}}
    \label{fig:pruning_res}
\end{figure*}

To evaluate our sparsity-driven pruning strategy, we compare DSP-LITETime with statically pruned models that use fixed filter counts per layer. We test static configurations with 16, 8 and 4 filters per layer, corresponding to pruning ratios of 55.8\%, 76.5\% and 84.9\% respectively.
Figure~\ref{fig:pruning_res} shows the results using MCM. 
Statistical tests show no significant accuracy difference between DSP-LITETime and LITETime\_16 model with the similar pruning ratio.
However, more aggressive static pruning (LITETime\_8 and LITETime\_4) leads to notable accuracy drops and statistically significant differences compared to DSP-LITETime. 
This shows that fixed pruning ratios struggle to adapt to different dataset complexities while our dynamic approach better maintains performance.

\begin{figure}[h]
    \centering
    \includegraphics[width=\linewidth]{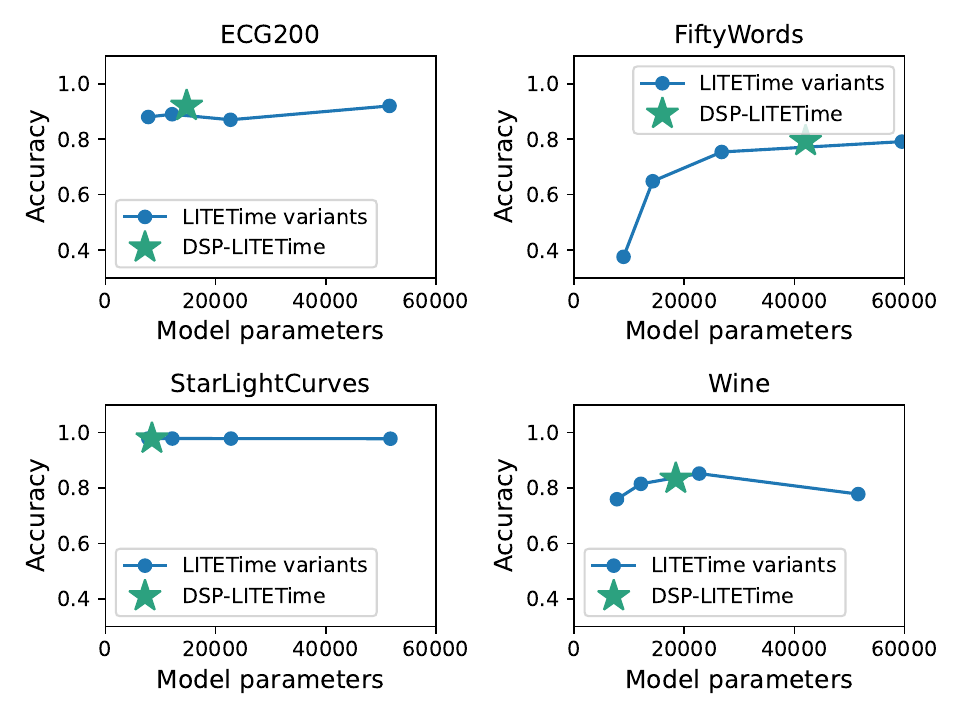}
    \caption{ Efficiency comparison of DSP-LITETime against the original LITETime and three static pruning variants of LITETime.}
    \label{fig:dynami_vs_static_pruning}
\end{figure}

Figure~\ref{fig:dynami_vs_static_pruning} shows how DSP-LITETime compares to the original LITETime and three static pruning versions (LITETime\_16, LITETime\_8, and LITETime\_4) on four datasets: \textit{ECG200}, \textit{FiftyWords}, \textit{StarLightCurves} and \textit{Wine}.
From the results we can observe that DSP-LITETime consistently achieves high accuracy while balancing performance and model size better than the static pruning methods.

Unlike static pruning which applies a fixed compression level and often requires extensive tuning, our dynamic approach adapts compression based on dataset complexity. 
This adaptivity enables a better trade-off between efficiency and accuracy, making DSP well-suited for large-scale or resource-constrained TSC scenarios.

\subsection{Redundancy Analysis}
\subsubsection{Quantitative Redundancy Analysis}
One of the reasons why our DSP approach achieves higher pruning ratios on both architectures is that these models inherently include a certain amount of redundancy in their learned representations. 
To validate this hypothesis, we analyze the redundancy present in the learned feature maps quantitatively. 
By assessing the effective rank of these feature maps, we can quantify the extent of redundancy and better understand how much of the model complexity is driven by redundant features. 

To measure feature redundancy in the learned representations, we compute the \textit{effective rank} of feature maps using Singular Value Decomposition (SVD)~\cite{eckart1936approximation}. 
For each training sample, the feature map at a given layer is treated as a matrix of shape $(C, T)$ where $C$ and $T$ denote the number of channels and temporal dimension respectively. 
We apply SVD to obtain the singular values and determine the smallest number of components needed to explain 95\% of the total variance. 
This number defines the \textit{effective rank}, a widely used measure of intrinsic dimensionality and redundancy~\cite{sedghi2018singular}.

However, since the pruned model contains fewer features compared to the base model, we introduce an \textit{efficiency score} (Equation~\ref{eq:efficiency_score}), calculated as the ratio between the effective rank and the number of activated channels:
\begin{equation}
\text{Efficiency Score} = \frac{\text{Effective Rank}}{\text{Number of Activated Channels}}
\label{eq:efficiency_score}
\end{equation}
A channel is considered active if its average activation over time is higher than the average activation of all channels in the same feature map.
This relative threshold allows us to identify channels that contribute more strongly compared to the overall activity and classifies them as active.
The effective rank is computed only on the activated channels of each feature map. 
This normalization enables a fair comparison across models with varying feature sizes.
A lower efficiency score indicates higher redundancy among the active channels while a score closer to 1 implies that each active channel contributes uniquely.

\begin{figure}
    \centering
    \includegraphics[width=0.75\linewidth]{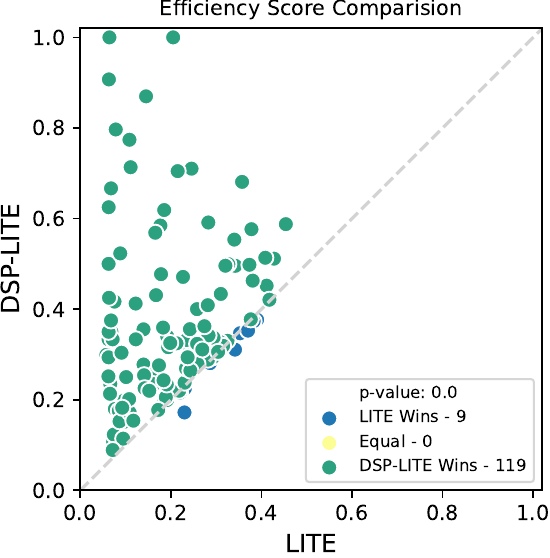}
    \caption{Efficiency score comparison across 128 UCR datasets. The pruned model demonstrates higher efficiency, reflecting reduced redundancy and more compact feature representations.}
    \label{fig:eff_score_comp}
\end{figure}

Figure~\ref{fig:eff_score_comp} presents the efficiency score comparison across 128 datasets from the UCR archive.
The results show that the base model consistently yields lower efficiency scores than the pruned model. 
These findings quantitatively support the argument that our sparsity-inducing training strategy not only reduces the number of active channels but also minimizes redundancy within them, leading to a more compact and efficient feature space.

\begin{figure*}
    \centering
    \includegraphics[width=0.7\linewidth]{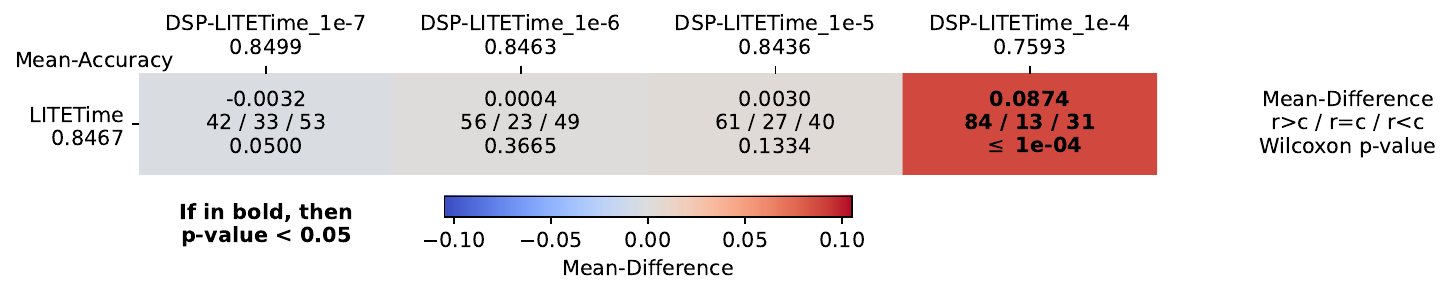}
    \caption{\centering{Performance comparison for different $\lambda$ values.}}
    \label{fig:lambda_mcm}
\end{figure*}

\subsubsection{Qualitative Redundancy Analysis}
\begin{figure}
    \centering
    \begin{subfigure}[b]{\linewidth}
        \includegraphics[width=\linewidth]{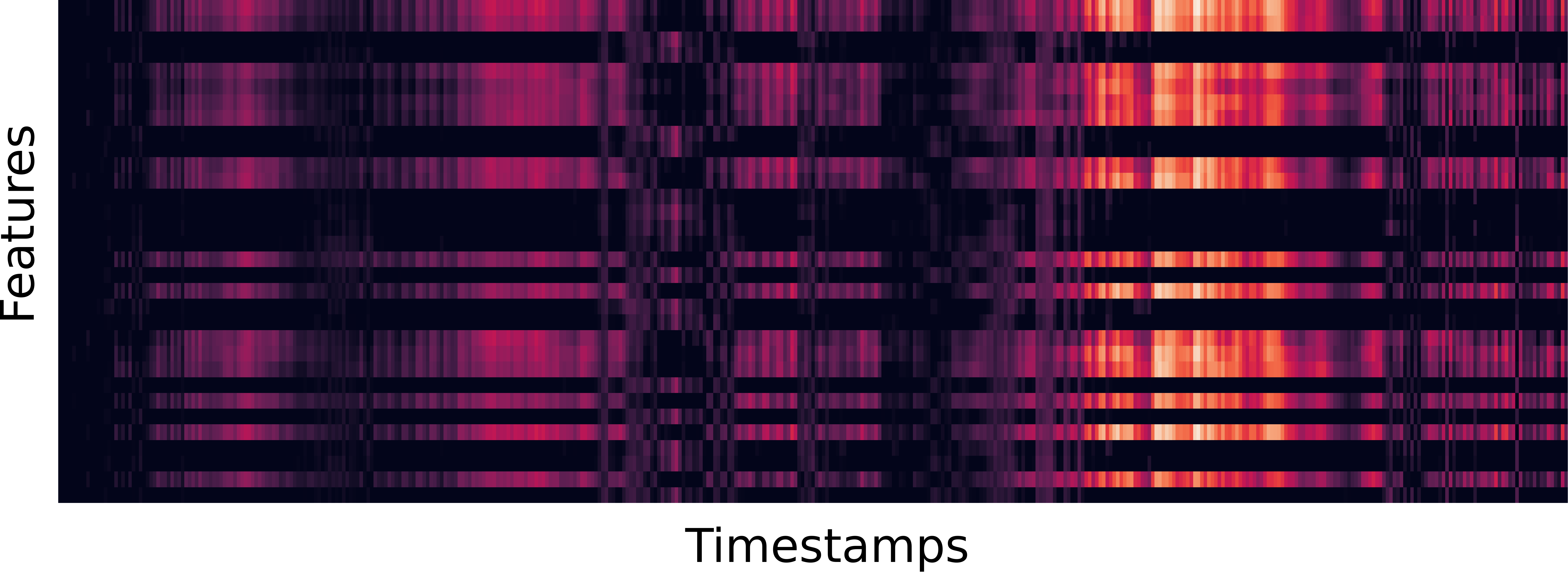}
        \caption{\centering{Classic LITE}}
        \label{fig:featuremap_base}
    \end{subfigure}
    
    \vspace{\floatsep} 
    
    \begin{subfigure}[b]{\linewidth}
        \includegraphics[width=\linewidth]{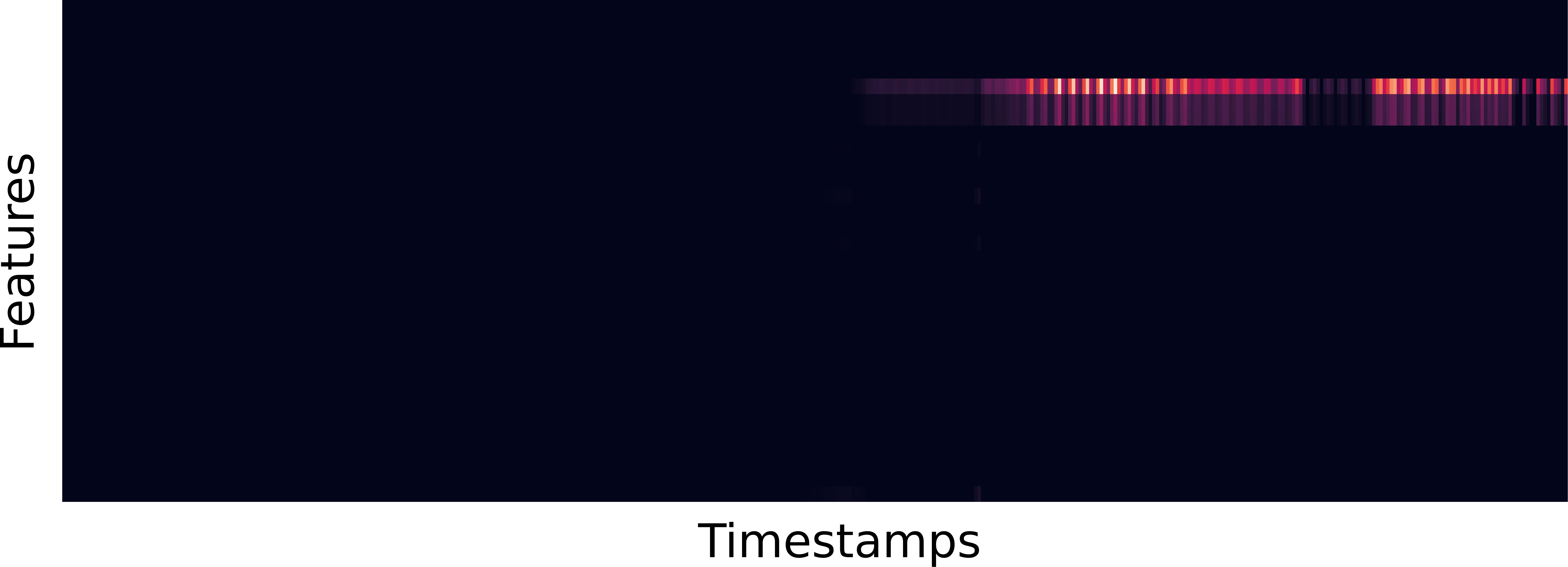}
        \caption{\centering{LITE model pretrained with sparsity loss}}
        \label{fig:featuremap_sparse}
    \end{subfigure}
    
    \caption{{Feature maps from the classic and pretrained LITE models illustrating redundancy and sparse activations.}}
    \label{fig:base_sparse_features}
\end{figure}

While the previous subsection quantitatively demonstrated redundancy differences using SVD, we now complement this analysis with a qualitative examination of the learned representations. 
Figure~\ref{fig:featuremap_base} and Figure~\ref{fig:featuremap_sparse} show feature maps from the base and pretrained LITE models (sparsity loss) respectively. 
In the classic LITE model (Figure~\ref{fig:featuremap_base}), we observe substantial redundancy where many channels activate on similar patterns, indicating repeated or overlapping information across features. 
In contrast, the pretrained LITE model (Figure~\ref{fig:featuremap_sparse}) exhibits a much more compact representation with only a few channels strongly activated.  
This visual distinction underscores the impact of the sparsity constraint in eliminating redundant features and retaining only the most discriminative ones. 
These qualitative observations align with the earlier quantitative results, further validating the effectiveness of DSP in promoting compact internal representations.

\subsection{Ablation study}
We tested different values of $\lambda$ on DSP-LITETime which controls impact of the sparsity loss as defined in Equation~\ref{eq:train_loss}. 
The values ranged from $0$ to $1e^{-4}$ and results are summarized in Figure~\ref{fig:lambda_mcm}. 
When $\lambda = 0$, it is equivalent to baseline.
The best average accuracy (0.8499) was achieved with $\lambda = 1e^{-7}$.
When $\lambda$ increased to $1e^{-6}$ and $1e^{-5}$ then average accuracy dropped slightly to 0.8463 and 0.8436 but remained close to LITETime without significant difference. 
However a large $\lambda = 1e^{-4}$ caused the average accuracy to fall sharply to 0.7593.
Statistical tests confirm that feature regularization reduces ability of the model to learn thus bigger  $\lambda$ values cause lower accuracy.
We selected $\lambda = 1e^{-5}$ in our experiments for both architectures as it balances pruning and accuracy well.

\section{Conclusion}\label{sec:conclusion}
This paper introduced Dynamic Structured Pruning (DSP), a fully automatic framework for structured pruning of CNNs in TSC. By integrating instance-wise sparsity loss during training with a global activation-based pruning strategy, DSP adaptively identifies and removes redundant filters. 
Comprehensive experiments on 128 diverse datasets from the UCR archive demonstrated that DSP achieves significant model compression with average pruning ratios of 58\% for LITETime and 75\% for InceptionTime while maintaining state-of-the-art classification accuracy.
Comparisons with static pruning methods and baselines highlighted the superior balance between accuracy preservation and compression offered by our approach.
Overall, DSP offers a scalable and practical solution for deploying efficient deep learning models on resource-constrained devices. 
In this work, DSP validated only on CNN architectures, extending it to other architectures is left for future research.
Future work also includes combining it with complementary compression methods such as quantization and knowledge distillation to further enhance model efficiency.

\bibliographystyle{apalike}
{\small
\bibliography{example}}

\end{document}